\documentclass{article}

\usepackage[preprint]{neurips_2022}

\usepackage[utf8]{inputenc} 
\usepackage[T1]{fontenc}    
\usepackage{hyperref}       
\usepackage{url}            
\usepackage{booktabs}       
\usepackage{amsfonts}       
\usepackage{nicefrac}       
\usepackage{microtype}      
\usepackage{xcolor}         
\usepackage{amsmath}
\usepackage{todonotes}
\usepackage[toc,page]{appendix}
\usepackage{multirow}
\usepackage{mathtools}

\title{LED: Latent Variable-based Estimation of Density}

\author{%
  Omri Ben-Dov \thanks{Max Planck Insitute of Intelligent Systems} \\
  \texttt{omri.ben-dov@tuebingen.mpg.de} \\
   \And
   Pravir Singh Gupta \\
   Mythic Inc. \\
   \texttt{pravir.singh.gupta@gmail.com} \\
   \AND
   Victoria Fernandez Abrevaya$ ^*$ \\
   \texttt{victoria.abrevaya@tuebingen.mpg.de} \\
   \And
   Michael J. Black$ ^*$\\
   \texttt{black@tuebingen.mpg.de} \\
   \And
   Partha Ghosh$ ^*$ \\
   \texttt{partha.ghosh@tuebingen.mpg.de} \\
}

\begin{document}

\maketitle

\newcommand{\rv}[1]{{\MakeUppercase{#1}}}  
\newcommand{\set}[1]{\mathcal{{\MakeUppercase{#1}}}}  
\newcommand{\vecspace}[1]{\mathbb{{\MakeUppercase{#1}}}} 

\newcommand{\argmax}{\operatornamewithlimits{argmax}} 
\newcommand{\argmin}{\operatornamewithlimits{argmin}} 
\newcommand{\nn}{\ensuremath{N_{\theta}}}

\newcommand{\FLOW}{FLOW }
\newcommand{\flow}{\FLOW}
\newcommand{\modelname}{LED}

\newcommand{\etal}{\textit{et al}. }
\newcommand{\eg}{e.g.,\ }
\newcommand{\ie}{i.e.,\ }
\newcommand{\etc}{etc.}
\newcommand{\qheading}[1]{\noindent\textbf{#1}}
\newcommand{\Qheading}[1]{\qheading{#1}}

\newcommand{\eqationref}[1]{Eq.~{\ref{#1}}}  
\newcommand{\figref}[1]{figure~{\ref{#1}}}  
\newcommand{\tableref}[1]{table~{\ref{#1}}}  
\newcommand{\secref}[1]{Section~{\ref{#1}}}  
\newcommand{\chapref}[1]{chapter~{\ref{#1}}}  

\newcommand{\todoinline}[2]{\textcolor{red}{TODO\{\MakeUppercase{#1}\}:} \textcolor{blue}{#2}}

\newcommand{\lossreal}{\mathcal{L}_{real}}
\newcommand{\lreal}{\lossreal}

\newcommand{\lossgen}{\mathcal{L}_{fake}}
\newcommand{\lgen}{\lossgen}

\begin{abstract}

Modern generative models are roughly divided into two main categories: (1) models that can produce high-quality random samples, but cannot estimate the exact density of new data points and (2) those that provide exact density estimation,
at the expense of sample quality and compactness of the latent space. 
In this work we propose LED, a new generative model closely related to GANs, that allows not only efficient sampling but also efficient density estimation. 
By maximizing log-likelihood on the output of the discriminator, we arrive at an alternative adversarial optimization objective that encourages generated data diversity. 
This formulation provides insights into the
relationships between several popular generative models.
Additionally, we construct a flow-based generator that can compute exact probabilities for generated samples, while allowing low-dimensional latent variables as input. 
Our experimental results, on various datasets, show that our density estimator produces accurate estimates, while retaining good quality in the generated samples.

  
\end{abstract}
\section{Introduction}
\label{sec:intro}



Generative models strive to extract some notion of the data distribution given a set of training data, either explicitly through a probability density (\eg~\cite{van2016pixel}), indirectly through a stochastic sampling mechanism (\eg~\cite{Goodfellow2014_GAN_GAN}), or both (\eg~\cite{dinh2014nice}). 
The current state of the art 
for generative modeling -- GANs \cite{Goodfellow2014_GAN_GAN} -- can handle large-dimensional data such as images with impressive performance, but they have only an \emph{implicit} notion of the distribution: they can generate random samples, but cannot compute the likelihood of a new data point. 
%
Access to an explicit density function, however, provides several advantages: quantitative comparison among models becomes straightforward; training through maximum likelihood estimation (MLE) has been proven to be asymptotically statistically efficient \cite{Huber67}; and applications such as unsupervised or semi-supervised learning can benefit from this prior knowledge. 

Autoregressive models \cite{van2016conditional, van2016pixel} and normalizing flows \cite{dinh2014nice} are among the most prominent examples of deep generative models that compute exact probability and employ direct log-likelihood maximization of their training dataset. 
 However, it is inefficient to sample from autoregressive models, and they do not provide a latent representation of the data. 
 Normalizing flows, on the other hand, allow both efficient sampling and density estimation, but make restrictive assumptions on the model architecture, requiring the latent space to be of the same dimensionality as the input, making it computationally expensive to use them in the 
 high-dimensional data regime.

Energy-Based models (EBMs)\cite{song2021train} and Variational Autoencoders (VAEs)~\cite{Kingma2014} are also deep generative models based on likelihood maximization. However, VAEs can only compute a lower bound of the likelihood and imposes restrictive assumptions on the family of distributions it can represent, resulting in inter-sample averaging. Consequently, when applied to images, VAEs generate blurry samples. EBMs, on the other hand, represent an unnormalized density, allowing greater flexibility in the choice of the functional form, at the cost of inefficient sampling and likelihood estimation. 


In this work, we derive a generative model that is both efficient at sampling and provides fast and explicit density estimates.
We start from a parametric model represented by a neural network $\nn$, and show how maximizing log-likelihood, conditioned on the network's output being a valid density, leads to the Wasserstein GAN~\cite{wasserstein_gan} formulation with only minor differences. 
This results in an adversarial training approach whose final product is both an efficient sampling method, on the generator side, and a density estimator on the discriminator side.
%
In doing so, we additionally propose a new architecture for the generator that combines flow models and adversarial learning. 
The combination of a GAN model with a flow-based generator has been explored before \cite{Grover18}. 
In contrast with previous work, however, our assumptions for MLE allows us to relax the requirements on the generator architecture (see \secref{sec:non_invertible_flow}).
With this, we construct a flow that performs up-sampling and down-sampling operations, starting from a lower-dimensional latent variable, where density estimation of real 
(as opposed to generated)
data points is relegated to the discriminator. 

Our experimental results on sample quality, test set encoding efficiency and encoding efficiency of generated samples, show that both our new generator, and the discriminator network $\nn$, work in harmony and are both of high quality (see \secref{LED:sec:expt}).

In summary, our contributions are i) Our method yields a model that can compute the normalized probability density function efficiently; ii) we relax the bijectivity constraint required by traditional \FLOW models, enabling more flexible generator architectures; 
iii) our optimization objective is closely related to that of GANs and our theoretical contributions clarify a hidden assumption made by the GAN objective.  Importantly, our work provides new insight into the properties of the discriminator, which can be regarded as a density estimator with minor changes. Furthermore, it connects GANs, normalizing \FLOW and EBMs from a theoretical perspective.

\section{Method}
\label{sec:method}

Given a dataset of i.i.d samples $\set{x} := \{x_i \in \vecspace{R}^n\}_{i=1}^{m}$, drawn from an unknown probability density $P_{data}$, the goal is to learn a parametric model $\nn$, with parameters $\theta \in \mathcal{M}$, that matches $P_{data}$. 
Here, we represent $\nn$ with a neural network. To make it a valid density function, we must ensure that $\nn$ is positive everywhere, and that it encloses unit mass. Henceforward, we call this normalized version of $\nn$ as the model density $P_\theta$ and the neural network embodying $\nn$ as the discriminator network (or the density estimator network) because of its similarity to the discriminator of a GAN. 
We determine the parameters $\theta$ by maximizing the expected log-likelihood, yielding the following constrained optimization:
\begin{equation}
    \argmax_{\theta \in \mathcal{M}}{\sum_{x_i}log(\nn(x_i))} \quad s.t. \quad \nn(x) > 0 \;\forall x \; ;\int_{\vecspace{R}^n} \nn(x)dx = 1 
    \label{log_likelihood_cop}
\end{equation}

It is easy to ensure a positive density by setting the last activation of the neural network to an exponential function. However, the integral on the right-hand side is generally intractable and hard to compute. We can approximate this integral by using Monte Carlo sampling, and rewrite it as follows:
%
%
%
%
%
 %
 \begin{equation}
    \alpha_\theta = \int_{\vecspace{R}^n}\nn(x)dx = \int_{\vecspace{R}^n}P_{G}(y)\frac{\nn(y)}{P_G(y)}dy = \mathbb{E}_{y\sim P_G(\rv{y})}\frac{\nn(y)}{P_G(y)}
    \label{int_as_expectation}
\end{equation}
 
 Here, $P_G(y)$ is an arbitrary density of the random variable $\rv{y} \in \vecspace{R}^n$ that is non-zero wherever the true data density $P(\rv{x})$ is non-zero and $\mathbb{E}$ represents the expectation operation. We have also assigned the shorthand $\alpha_\theta$ for future reference to this integral. This, however, is analytically intractable. Therefore, we use a finite sample approximation of the expectation.

The problem with this method, however, is that the sample complexity is unrealistically large when $y\in \vecspace{R}^n$ and $n$ is large. The sample efficiency can be greatly boosted by performing importance sampling, especially so, when the true data density is highly concentrated. This is widely believed to be the case with natural images, since neighboring pixels are highly correlated. Therefore, we propose the following technique to efficiently get samples $y_j = G_\phi(z_j, \eta_j)$, to compute the expectation. Here, $z_j \in \vecspace{R}^d$ is a latent vector with $d<<n$, $G_\phi$ is the generator, which is a deterministic function that embeds the latent variable into the data space, and $\eta_j \in \vecspace{R}^{n-d}$ is auxiliary noise. We further assume a prior distribution $P(\rv{z})$ for the latent variable $z_j$ and an independent isotopic Gaussian distribution with variance $\sigma*I$, for the auxiliary noise $\eta_j$. Here, $I$ denotes an identity matrix of appropriate size. This lets us efficiently approximate the expectation in \eqationref{int_as_expectation} with $\sum_{z_j\sim P(\rv{z}), \eta_j\sim\mathcal{N}(0, \sigma*I)}[\nn(G_\phi(z_j, \eta_j))/P_G(\rv{y}=G_\phi(z_j, \eta_j))]$. Given this, \eqationref{log_likelihood_cop} can now be rewritten as in \eqationref{log_likelihood_cop_with_generator}
\begin{equation}
    \begin{split}
        & \argmax_\theta{\sum_{x_i}log\left(\frac{\nn(x_i)}{\frac{1}{N}\sum_{z_j, \eta_j}\nn(G_\phi(z_j, \eta_j))/P_G(G_\phi(z_j, \eta_j))}\right)} \\ & = \argmax_\theta{\sum_{x_i}\left(log(\nn(x_i)) - log\left(\sum_{z_j, \eta_j}\frac{\nn(G_\phi(z_j, \eta_j))}{P_G(G_\phi(z_j, \eta_j))}\right)\right)} = \lreal + \lgen
    \end{split}
    \label{log_likelihood_cop_with_generator}
\end{equation}

\eqationref{log_likelihood_cop_with_generator} serves as our objective function for training $\theta$. Here, we define $\lreal := log(\nn(x_i))$ and $\lgen := -log(\sum_{z_j, \eta_j}\nn(G_\phi(z_j, \eta_j))/P_G(G_\phi(z_j, \eta_j)))$ for quick reference. Note that \eqationref{log_likelihood_cop_with_generator} is the same objective function as in EBMs, except that here we explicitly define the integral, or its approximation.
If we approximate $\lgen$ by a single sample of $Z_j$ and $\eta_j$, then \eqationref{log_likelihood_cop_with_generator} can further be simplified, as in \eqationref{gan_obj_recovered}. Further, $log\, P_G(G_\phi(z_j, \eta_j))$ is not dependent on  $\theta$ and hence can be dropped from this optimization. This fully recovers the usual GAN objective for the critic network.
\begin{equation}
    \begin{split}
        & \argmax_\theta{\sum_{x_i}\left(log(\nn(x_i)) - log\left(\sum_{z^i}c_z^i\nn(G_\phi(z_j, \eta_j))\right)\right)} \\ &
        \approx \argmax_\theta{\sum_{x_i}\left(log(\nn(x_i)) - log\left(\nn(G_\phi(z_j, \eta_j))\right) - log(P_G(G_\phi(z_j, \eta_j)))\right)}
    \end{split}
    \label{gan_obj_recovered}
\end{equation}

This equation, however, requires a suitable function $G_\phi$. We choose to use a neural network to represent this function, and let us call this network the generator network. To compute $\lgen$, clearly one needs to have access to $P_G(\rv{y})$ evaluated at $y_j = G_\phi(z_j, \eta_j)$ for arbitrary $z_j$ and $\eta_j$. We could in principle obtain this by using normalizing flows as the generator, but this would impose strong restrictions on the architecture.
Specifically, this would require the whole network to be analytically invertible, and consequently the latent space cannot be of lower dimensionality than the input data. 
However, note that in our case we can relax this requirement, since we only need to evaluate $P_G(\rv{y})$ at points that have the form $y_j = G_\phi(z_j, \eta_j)$, \ie only at generated points.
We discuss how to construct such a flow 
in \secref{sec:non_invertible_flow}.

\subsection{Generator objective}
\label{LED:gen_obj}
Before diving deep in to the details of the generator architecture, it is worth examining the objective function with which we intend to optimize the parameters of it. A quick inspection of \eqationref{int_as_expectation} shows that it has the form of the importance sampling mechanism. In such scenarios, often the biased distribution ($P_G(\rv{y})$ in our case) is chosen such that the variance of the weighted samples is reduced \cite{rubinstein2016simulation}. In our case, the variance of $\frac{\nn(y)}{P_G(y)}$ is minimized if we match ${\nn(y)}$ and $P_G(y)$ up to a multiplicative factor, more precisely $P_G(y) = \nn(y)/\alpha_\theta$, everywhere. Here, $\alpha$ is the integral constant and is not known precisely. We do so by minimizing a divergence measure between $P_G(\rv{y})$ and $\nn(\rv{y})$. Here, $\nn(\rv{y})$ as usual, represents the unnormalized density that we seek to fit to our training data. We choose the KL divergence measure \cite{kullback1951information}. We do so specifically because, as seen from \eqationref{kl_rejects_normalization}, the optimization for the generator parameter becomes independent of the normalizing constant of the distribution given by $\nn$.
\begin{equation}
    \begin{split}
        & \argmin_\phi{\mathbb{E}_{y_i \sim P_G(\rv{y})} log\left(\frac{P_G(y_i)\alpha_\theta}{\nn(y_i)}\right)} = \argmax_\phi{\mathbb{E}_{y_i \sim P_G(\rv{y})}[-log P_G(y_i) + log \nn(y_i)} - log \alpha_\theta]\\
        & = \argmax_\phi\left[ H(P_G(\rv{y})) + \mathbb{E}_{z_i \sim P(\rv{z})} log \nn(G_\phi(z_i))\right]
    \end{split}
    \label{kl_rejects_normalization}
\end{equation}
Here $H(P_G(\rv{y}))$ represents the entropy of the generator distribution. A close look at \eqationref{kl_rejects_normalization} reveals that it is identical to the generator objective of a Wasserstein-GAN \cite{wasserstein_gan} except for the additional entropy term. This term precisely seeks to increase diversity of the generated samples. 

\subsubsection{Generator function using a non-invertible flow}
\label{sec:non_invertible_flow}
To take advantage of the above-mentioned relaxed requirement of our generator function $G_\phi$ we propose a variant of normalizing-flow networks. In fact, in this type of network we neither maintain dimensionality nor invertibility, therefore strictly speaking they are not flow networks. These networks are composed of layers of simple functions that operate on a random variable to produce a transformed random variable, such that given the density of the input random variable, we can efficiently compute the density of the output random variable. For this purpose, we borrow the change of variable mechanism, formally described in \eqationref{trck_probabilities_inact}. The difference between our network and traditional normalizing-flow networks is that our model neither needs to be efficiently invertible, nor does it always preserve dimensionality. We can both increase and decrease dimensionality. This ability precisely let us build architectures close to modern GAN architectures, which have heuristically been found to work well. Note that, in exchange, we cannot compute probabilities for random images -- only for generated ones. In our framework, this function is provided by the discriminator instead. 

\subsubsection{Increasing dimensionality}
\label{sec:led:change_of_variable}
Given a function $t = g(u) : \vecspace{R}^o \rightarrow \vecspace{R}^o$, the change of variable formula dictates that $P(\rv{t}) = P(\rv{u})\left| \frac{\partial g}{\partial u}\right|^{-1}$. A multi-layer neural network, \eg our generator $G$ with $l$ layers, can be thought of as a series of operations; \ie $ G := g_1(g_2(g_3( ... g_l(.))))$, formally, $t_1 = g_1(u_1), t_2 = g_2(u_2), \;...\; t_l = g_l(u_l)$, and $t_2 = u_1, t_3 = u_2, \;...\; t_l = u_{l-1}$. In such a scenario, if $t_1, t_2, t_3 \;...\; t_l, u_1, u_2, u_3, \;...\;u_l \in \vecspace{R}^o $, one can track the density of the activations as follows in \eqationref{trck_probabilities_inact}
\begin{equation}
    P(t_k) = P(u_k)\left|\frac{\partial g_k}{\partial u_k}\right|^{-1} \; \forall k \in \{1 \;...\; l\}
    \label{trck_probabilities_inact}
\end{equation} 

A sufficient criterion for this to remain valid is bijectivity of, $g_k \forall k$ and that, in turn, enforces the dimensionality of $z_k$ and $a_k$ to be equal. However, we are free to redefine $t_k := [u_{k-1} || an_k]$. Here, `||' represents a concatenation operation and $\eta_j \sim \mathcal{N}(0, I)$, where $\mathcal{N}$ is the normal distribution with mean zero and identity variance. This enables us to introduce noise at multiple layers of operation in the generator and, starting from a low-dimensional latent space, generate high-dimensional data, similar to StyleGAN \cite{karras2019style}. This modification and \eqationref{trck_probabilities_inact} enables us to write \eqationref{trck_probabilities_inact_noise}
\begin{equation}
    P(t_{k+1}) = P(t_k)\left|\frac{\partial g_k}{\partial t_k}\right|^{-1}P(an_k)  \;\; \forall k \in \{1 \;...\; l\}
    \label{trck_probabilities_inact_noise}
\end{equation}

\subsubsection{Dimensionality reduction}
\label{LED:sec:dim_red}
To closely follow traditional generator architecture, \eg the DC-GAN architecture \cite{radford2015dcgan}, one necessary operation is dimensionality reduction, \ie to have operations of the form $t_k = g_k(u_k) : \vecspace{R}^{o_1} \rightarrow \vecspace{R}^{o_2} \; ; o_1 > o_2$. Moreover, to compute $\lgen$, given the density of the input random variable $P(\rv{u}_k = u_k)$, one must be able to efficiently compute the density of the output random variable $P(\rv{t}_k = g_k(u_k))$. 
For the time being, imagine that ${t}_k \in \vecspace{R}^{o_2}$, \ie $g_k$ is a dimensionality preserving operation. Now, let us group $t_k$ in two parts, \ie $t_k = [t_k^1|| t_k^2]$. Here, `||' represents the concatenation operation and $t_k^1 \in \vecspace{R}^{o_2}, t_k^2 \in \vecspace{R}^{o_1- o_2}$. In such a setting, a dimensionality reduction operation would simply ignore $t_k^2$ and recover a function between $\vecspace{R}^{o_1} \rightarrow \vecspace{R}^{o_2}$. Any such function, however, requires us to compute $P(\rv{t}^1_k=t_k^1)$ given $P(\rv{t}_k=t_k):=P(\rv{t}_k=[t^1_k|t_k^2]):=P(\rv{T}_k^1, \rv{T}_k^2)$. This is a marginalization operation and can be computed as $\int_{\rv{t}^2_k}P(\rv{T}_k^1, \rv{T}_k^2)d\rv{T}_k^2$. This integral is unfortunately intractable in the general case. Therefore, we make a case-specific assumption, namely, we assume that, in real images, a pixel is normally distributed given its immediate neighbor. This assumption lets us marginalize $\rv{T}_k^2$ efficiently, as shown in \eqationref{trck_probabilities_dim_reduction}
\begin{equation}
    P(\rv{T}_k^1) = \frac{P(\rv{T}_k^1, \rv{T}_k^2)}{P(\rv{T}_k^2 | \rv{T}_k^1)} = \frac{P(\rv{T}_k^1, \rv{T}_k^2)}{\mathcal{N}(\rv{T}_k^1, \sigma*I)}
    \label{trck_probabilities_dim_reduction}
\end{equation}

Here in \eqationref{trck_probabilities_dim_reduction}, we get the second equality using our assumption stated above, $\sigma$ represents a tunable variance of the normal distribution, and $I$ is an identity matrix.

\subsection{Tailoring the candidate density function}
\label{sec:perp_cmp_in_prob_comp}
We chose to represent $\nn$, with a deep neural network. A deep neural network is a composition of repeated linear projection operations with nonlinear activations. Furthermore, often the linear operations, \eg strided convolution, reduce dimensions by projecting the input vector to a subspace. These operations hope to retain the most `useful' dimensions for the task at hand. In our case, the network is encouraged to produce high responses for points coming from a dataset. Therefore, the linear layers are expected to align themselves to the axis of the highest data variance. Now, if we use this network for points that do not come from our training data set or are not even are of the same kind, \eg non images, they are expected to have significant variance on axes different from the high-variance axes of the data. We can exploit this feature, to assign low probability to such samples. Concretely, a linear operation with weight matrix $w$ operating on a data vector $x$ can be expressed as a matrix vector multiplication $t_\parallel = w*x$. Here $t_\parallel$ is the component of $x$ in the subspace spanned by the rows \footnote{considering $x$ is a column vector.} of $w$. We can recover the component perpendicular to this subspace \footnote{assuming that $w$ is row rank deficient.}  $t_\bot = x - \frac{t_\parallel^T*w}{||t_\parallel^T*w||_2}*x$. Here, $\{\cdot\}^T$ represents the transpose operation and $||\cdot||_2$ represents the vector norm operation. As discussed above, we expect this perpendicular component $t_\bot$ to be large for points that do not belong to our training data and therefore should be assigned low density. Therefore our density network is evaluated as $\nn(x)e^{-\sum_k||t_{\bot k}||_2}$. Here $t_{\bot k}$ represents the perpendicular component at every linear projection layer. 

\subsection{Architecture}

LED consists of a generator and a discriminator. The discriminator in our case plays the role of a density estimator. 
As described in \secref{LED:gen_obj}, the generator consists of $l$ transformations given by $g_k \; ; \; k\in\{1, 2, 3 ... l\}$. We choose these transformations carefully such that, given the density of the input variable, we can efficiently compute the density of the output variable. A traditional GAN architecture often employs fully connected, convolution, batch normalization, ReLU and interpolation layers. We implement a mechanism to keep track of the probability density through all of these operations; precise implementation details can be found in the supplementary material, \secref{LED:sec:architecture}. Our final generator architecture closely follows that of DCGAN~\cite{radford2015dcgan}, known to perform well with adversarial training.  
The discriminator architecture follows precisely that of DCGAN, except that we subtract the perpendicular component in the convolution layers (\secref{sec:perp_cmp_in_prob_comp}). 
\section{Experiments}
\label{LED:sec:expt}
In this section, we provide experimental results for the generated data and the density estimation using both synthetic (\secref{sec:results_synth_data}) and real (\secref{sec:results_real_data}) datasets. We show qualitative examples in \secref{sec:results_qual}. Implementation details can be found in the Sup. Mat., \secref{sec:implemetation_dets}.





\subsection{Synthetic Data} 
\label{sec:results_synth_data}

\begin{figure}
    \centering
    \includegraphics[width=\columnwidth]{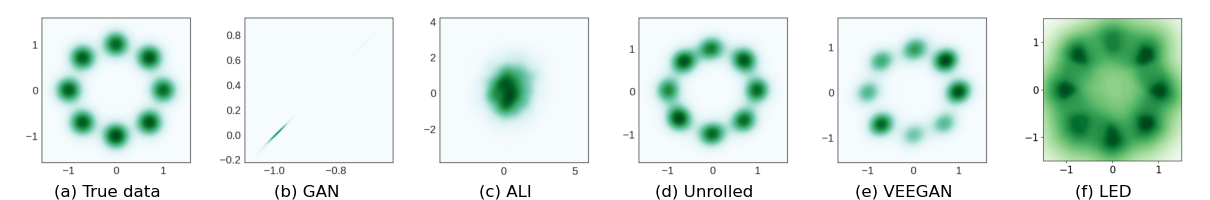}
    \includegraphics[width=\columnwidth]{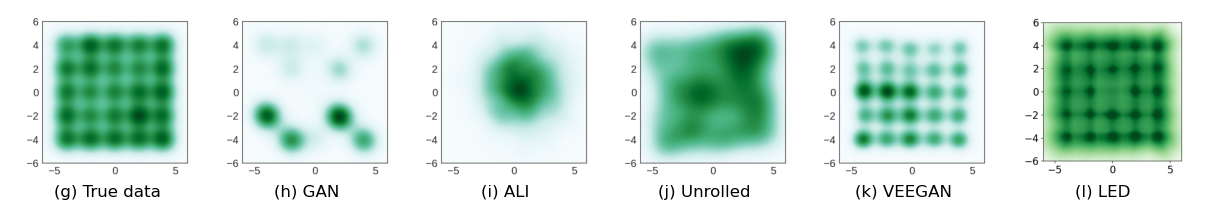}
    \caption{Density modeled by different generative models. Top row: Ring of Gaussians. Bottom row: Grid of Gaussians. (a-e,g-k) are taken from \cite{srivastava2017veegan} and were produced by kernel density estimation run on generated samples. (f,l) the direct density (in log scale) computed by LED. Note that our model assigns a non-zero probability to regions where the KDE does not, since only a finite number of samples were used to estimate generator density in the experiments (a-e,g-k).}
    \label{fig:toy_dists}
\end{figure}

We begin by examining the notion of data distribution captured by our \textbf{discriminator} model (\ie density estimator), compared to other GAN-based methods. 
To this end, we perform an experiment on a toy 2D dataset, similar to the one presented in VEEGAN~\cite{srivastava2017veegan}. 
%
In particular, we train our model on two sets of Gaussian Mixture Models (GMM), with one set comprising $8$ modes forming a ring (Fig. \ref{fig:toy_dists}a) and another set comprising $25$ modes in a grid (Fig. \ref{fig:toy_dists}g). 

Since GAN models normally do not return a direct estimate of the probability of the data, the compared methods use Kernel Density Estimation (KDE) to visualize the relevant probability, as shown in Fig. \ref{fig:toy_dists}(a-e, g-k). Note that LED instead has direct access to this value through the discriminator. 
We observe that the density learned by LED (shown in Fig. \ref{fig:toy_dists}(a-e, g-k)) matches the true distribution, returning high probability for all modes, and low probability away from them. 



We also quantify the quality of the density captured by the \textbf{generator}. For this, we use the High-Quality Samples metric from \cite{srivastava2017veegan}, where a generated point is considered \emph{high quality} if it is within a $3\sigma$ distance from the nearest mode. We generate 2,500 points and report the percentage of points that are high quality over five runs. We can see here (in \tableref{table:2dgen_hq}) that our generator produces significantly higher quality samples than other models, achieving $10\%$ more than VEEGAN~\cite{srivastava2017veegan}. On the other hand, our model is not capable of fully capturing all $25$ modes in the second GMM. %
%
%
We hypothesize that this is due to an insufficient expressive power of the generator. 
To test this hypothesis, we follow this following experiment protocol. First, with an intentionally weakened generator, we capture a smaller number of modes. We find that this is consistent and reproducible. Next, when we progressively increase the power of the generator, we see that our model progressively generates more modes.
The result of this experiment, along with qualitative results for the generator density, can be found in the Sup. Mat., \secref{sec:gen_power_2d}.

 
 
\begin{table}[!ht]
    \centering
    \begin{tabular}{@{}lcccccc@{}}
\toprule
\multirow{2}{*}{}     & \multicolumn{2}{c}{\textbf{2D Ring}}   & \multicolumn{2}{c}{\textbf{2D Grid}}    \\  \cmidrule(r{1em}l{1em}){2-3} \cmidrule(r{1em}l{1em}){4-5}
                      & \textbf{\begin{tabular}[c]{@{}c@{}}Modes \\ 
                      (Max 8)\end{tabular}} & \textbf{\begin{tabular}[c]{@{}c@{}}\% HQ\\ 
                      Samples\end{tabular}} & \textbf{\begin{tabular}[c]{@{}c@{}}Modes \\ 
                      (Max 25)\end{tabular}} & \textbf{\begin{tabular}[c]{@{}c@{}}\% HQ\\
                      Samples\end{tabular}} \\ 
                      \midrule
\textbf{GAN \cite{Goodfellow2014_GAN_GAN}}          & 1        & 99.3            &3.3               &0.5           \\ 
\textbf{ALI \cite{DumoulinBPLAMC17ALI}}          & 2.8      & 0.13        & 15.8         & 1.6          \\ 
\textbf{Unrolled GAN \cite{MetzPPS17unrolled}} & 7.6      & 35.6    & 23.6        & 16            \\ 
\textbf{VEEGAN \cite{srivastava2017veegan}}     & \textbf{8}       & 52.9         & \textbf{24.6}         & 40               \\ 
\textbf{LED (ours)}     &  \textbf{8}       & \textbf{63.5}        & 16.2         & \textbf{68.3}               \\ 
\bottomrule
\end{tabular}

     \vspace{2.5mm}
     \caption{The high quality (HQ) samples percentages of various generators for 2D GMMs. The numbers for all models, except LED (ours), are taken from \cite{srivastava2017veegan}. LED produces higher quality samples than all other tested models. Here, high-quality samples are defined to be samples are within $3\sigma$ of a real data mode.}

    \label{table:2dgen_hq}
\end{table}
 

\subsection{Real Data}
\label{sec:results_real_data}

We now show results of LED over real data generation, where we compare against the 
flow-based model GLOW~\cite{Kingma2018_glow}, 
and a GAN baseline that uses our same architecture, but with a standard GAN loss~\cite{Kingma2014}.
We also compare the results of different numbers of samples for the integration approximation.
We train these models with the MNIST~\cite{mnist} and CelebA~\cite{celeba} datasets and evaluate the results using (1) Fréchet Inception Distance (FID), and (2) bits-per-pixel ($b/p$) of the test set.
We compute the $b/p$ of real images by computing their negative-log-likelihood using the score given by the discriminator along with \eqationref{int_as_expectation}, and subsequently normalizing the result by the number of pixels $D$ : $\textit{b/p}=\frac{1}{n}\sum_{x\in X}\frac{d\left(x\right)}{\alpha_{0}D\ln2}$.
We also compute the $b/p$ for the generated images using the discriminator score and the generator probability.
For the discriminator score, we use the same method as with the real images.
For the generator $b/p$ we use the change of variables formula, as we know the latent vector and its probability along with the log-determinant of the generator's Jacobean: $\ln P_g\left(G\left(z\right)\right)=\ln P\left(z\right)-\ln\left|\frac{\partial G\left(z\right)}{\partial z}\right|$.

\begin{figure}
    \centering
    \includegraphics[width=\columnwidth]{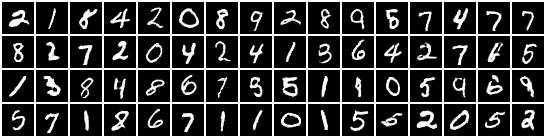}
    \caption{Random examples of generated images using MNIST at 32 × 32 resolution. FID 4.19.}
    \label{fig:mnist_res}
\end{figure}

\begin{figure}
    \centering
    \includegraphics[width=0.99\columnwidth]{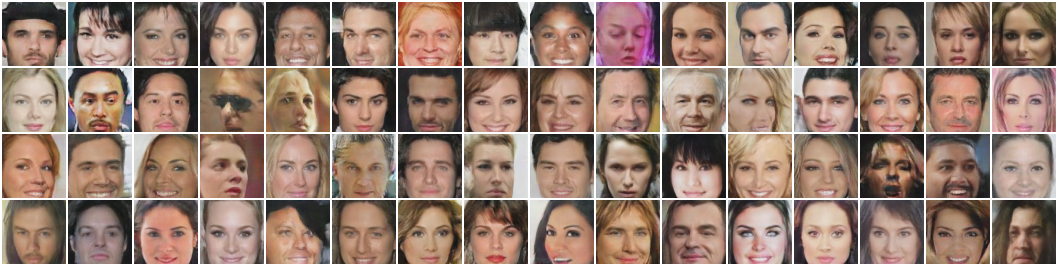}
    \caption{Random examples of generated images using CelebA at $64\times64$ resolution. FID $12.6$.}
    \label{fig:celeba_res}
\end{figure}

Generated samples can be seen in Fig. \ref{fig:mnist_res} for MNIST dataset and in Fig. \ref{fig:celeba_res} for CelebA dataset. The corresponding FID and $b/p$ results are reported in Tbl. \ref{table:FID_score_b_ps}.
LED outperforms the current state of the art in terms of $b/p$ metric, which increases with the number of samples used for the integral approximation. 
Looking at the FID scores, we see that using only one sample results in unrealistic images (high FID). This is understandable since this approach is equivalent to training a generator with only a single sample for every batch of the discriminator, which has been observed to perform poorly~\cite{Brock18}. 
Increasing the number of samples significantly improves the FID scores to comparative values with the state of the art.
Tbl. \ref{table:FID_score_b_ps} also shows that subtracting the perpendicular part (Sec. \ref{sec:perp_cmp_in_prob_comp}) in the discriminator (\secref{sec:perp_cmp_in_prob_comp}) results in a better quality image (lower FID) and better efficiency (lower $b/p$).

\subsection{Qualitative Results} 
\label{sec:results_qual}

\begin{figure}
    \centering
    \includegraphics[width=0.99\columnwidth]{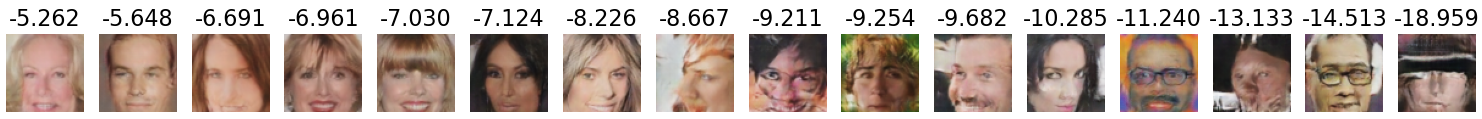}
    \caption{Random samples of generated images with their discriminator-assigned probability above. }
    \label{fig:short_prob_sort}
\end{figure}

In Fig. \ref{fig:short_prob_sort} (and in \secref{sec:sort_prob_res} in Sup. Mat.) we visually assess the probabilities assigned by the discriminator by showing random samples sorted by density. We can see here that high quality images get higher probabilities than low quality images, suggesting that our trained discriminator correctly captures the distribution of the given dataset.

We also demonstrate the power of the discriminator by optimizing over its score. 
We begin by generating images using latent vectors that are between $1.5$ and $3\sigma$ away from the origin, which are usually of low quality. 
We then optimize the latent vectors such that they maximize the output of the discriminator. 
The results, in the supplementary material, \secref{sec:mnsit_opt_far} show that this approach can significantly increase the quality of the images for MNIST.

\begin{figure}
    \centering
    \includegraphics[width=0.99\columnwidth]{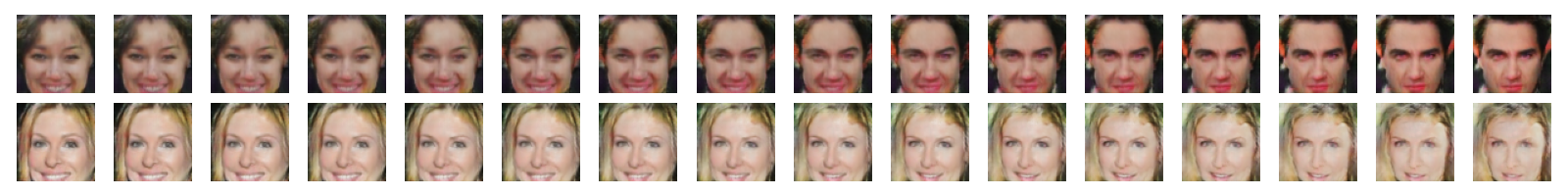}
    \caption{Generated images from linear interpolations of the latent space, using the CelebA dataset.}
    \label{fig:short_interp}
\end{figure}

Finally, we show in Fig. \ref{fig:short_interp} (and in \secref{sec:interp_mnist} in Sup. Mat.) qualitative results for the generator on an interpolation experiment. 


 \begin{table}[!ht]
    \centering
    \begin{tabular}{@{} c c c c c c c c c c c c c c c @{}}
        \toprule
            & \multicolumn{3}{c}{MNIST}  & \multicolumn{3}{c}{CelebA} \\
            \cmidrule(lr){2-4} \cmidrule(lr){5-7}   \\
            Model & FID & b/p & smpl & FID & b/p & smpl \\
                  &     &     & b/p  &     &     & b/p   \\
                \cmidrule(lr){2-4} \cmidrule(lr){5-7}   \\
            DC-GAN & 5.6 & -- & --/--  & (12.5)13 & -- & --/--  \\
            GLOW & 11 & -- & --/-- & -- & -- & --/--  \\
            Ours-1sa & 118 & 0.509 & 1.011 / 8.245  & 209 & 0.019 & 0.025/2.366  \\
            Ours-64sa & 4.3 & 0.546 & 0.992 / 8.75  & 24 & 0.009 & 0.012/1.049\\
            Ours-256sa & 4.19 & 0.530 & 1.003 / 8.515 & 36 & 0.022 & 0.022/0.012\\
        \bottomrule
    \end{tabular}

     \vspace{2.5mm}
     \caption{FID and bits-per-pixel ($b/p$) of the held-out test set, and $b/p$ of the generated samples. Note that the latter can be evaluated using our flow-based generator, as well as the independent density estimator network; we report them as \emph{generator $b/p$ / critic $b/p$}. 
    Here, 64sa, 256sa stands for the number of samples used to approximate the second sum of \eqationref{log_likelihood_cop_with_generator}. The GLOW $b/p$ and FID for CelebA could not be calculated since it failed to converge with default parameter settings.}
    \label{table:FID_score_b_ps}
\end{table}

\section{Related work}

Two main categories of generative models are \emph{prescribed} and \emph{implicit} models~\cite{Diggle84}. Prescribed models recover an explicit parametric specification of the density function and are trained and evaluated by maximum likelihood estimation (MLE); our work belongs to this family. 
%
Implicit models, on the other hand, represent the data distribution indirectly through a stochastic mechanism that generates random samples. In general, this offers more flexibility in terms of learning objective and model architecture, which is hypothesized to be responsible for the high visual quality of the generated samples. 

Normalizing flows \cite{kobyzev2020normalizing, dinh2016density, Kingma2018_glow} and autoregressive models \cite{theis2015generative, van2016pixel, salimans2017pixelcnn++} are examples of deep \emph{prescribed} generative models. 
Since these compute the density function explicitly, they can be optimized and evaluated using the train and test-set log-likelihood. Although autoregressive models can efficiently work with high-dimensional data during training, due to ancestral sampling they are extremely slow at generating new samples. Normalizing flows require an invertible architecture to compute the likelihood, and consequently can only  support latent spaces of the same dimensionality as the input data. In addition, they tend to yield large and memory-hungry models, and are therefore not so suitable for high-dimensional data. 
In this work, we relax the invertibility constraint by computing the flow in only one direction, enabling the use of lower-dimensional latent vectors, and more computational resource efficient architectures. 





Generative Adversarial Networks (GANs) are the most prominent example of \emph{implicit} models. GANs currently produce state-of-the-art generated sample quality~\cite{karras2019analyzing}. However, it has been observed that GANs may trade diversity for precision~\cite{Che16, Salimans16, srivastava2017veegan}. This results in generators that produce samples from only a few modes of the data distribution, a phenomenon known as `mode collapse'.
GANs are also well known for having unstable training dynamics 
\cite{wasserstein_gan, mescheder2018training, heusel2017gans}. 


An intermediate category of generative models considers only an \emph{approximation} to the density function.
Examples include a lower bound on the likelihood for VAEs~\cite{Kingma2014} and diffusion models~\cite{ho2020denoising}, or the unnormalized density in the case of EBMs~\cite{song2021train}. 
VAEs are known to suffer from low generation quality, \ie they tend to produce blurry samples. Diffusion models can generate images of very high sample-quality~\cite{dhariwal2021diffusion}; however, the latent representation needs to be of the same dimension as the input data. EBMs 
deploy several techniques to obtain 
the derivative of the normalizing factor with respect to the model parameters. We maximize the same cost function as EBMs (see \eqationref{log_likelihood_cop}), but explicitly model the normalization constant $\alpha_\theta$. 
As explained in \secref{sec:method}, this leads to a variant of the GAN formulation, thus making connections between the three models. 

\section{Conclusion}
\label{LED:sec:conclusion}
We presented a new generative model, \modelname, which is closely related to GANs while being able to efficiently compute the density function of the data. Our formulation, based on maximizing log-likelihood of the discriminator output, recovers a training objective that is similar to the objective function of GANs, but with
a crucial difference that introduces 
an entropy maximization term in the generator objective. Our experimental results show that LED  produces a generator that is on par with other GAN generators, along with accurate density estimations.
LED 
provides new understandings on the properties of the discriminator, and provides insights into GANs from a maximum likelihood perspective, while connecting these with EBMs. We believe this can open up promising directions into understanding and improving generative models. 

\clearpage
\bibliographystyle{apalike}
\bibliography{references}

\section*{Checklist}

The checklist follows the references.  Please
read the checklist guidelines carefully for information on how to answer these
questions.  For each question, change the default \answerTODO{} to \answerYes{},
\answerNo{}, or \answerNA{}.  You are strongly encouraged to include a {\bf
justification to your answer}, either by referencing the appropriate section of
your paper or providing a brief inline description.  For example:
\begin{itemize}
  \item Did you include the license to the code and datasets? \answerYes{In Sup. Mat.}
  \item Did you include the license to the code and datasets? \answerNo{The code and the data are proprietary.}
  \item Did you include the license to the code and datasets? \answerNA{}
\end{itemize}
Please do not modify the questions and only use the provided macros for your
answers.  Note that the Checklist section does not count towards the page
limit.  In your paper, please delete this instructions block and only keep the
Checklist section heading above along with the questions/answers below.

\begin{enumerate}

\item For all authors...
\begin{enumerate}
  \item Do the main claims made in the abstract and introduction accurately reflect the paper's contributions and scope?
    \answerYes{}
  \item Did you describe the limitations of your work?
    \answerNo{}
  \item Did you discuss any potential negative societal impacts of your work?
    \answerNo{}
  \item Have you read the ethics review guidelines and ensured that your paper conforms to them?
    \answerYes{}
\end{enumerate}

\item If you are including theoretical results...
\begin{enumerate}
  \item Did you state the full set of assumptions of all theoretical results?
    \answerYes{}
        \item Did you include complete proofs of all theoretical results?
    \answerYes{}
\end{enumerate}

\item If you ran experiments...
\begin{enumerate}
  \item Did you include the code, data, and instructions needed to reproduce the main experimental results (either in the supplemental material or as a URL)?
    \answerYes{}
  \item Did you specify all the training details (e.g., data splits, hyperparameters, how they were chosen)?
    \answerYes{}
        \item Did you report error bars (e.g., with respect to the random seed after running experiments multiple times)?
    \answerNo{}
        \item Did you include the total amount of compute and the type of resources used (e.g., type of GPUs, internal cluster, or cloud provider)?
    \answerNo{}
\end{enumerate}

\item If you are using existing assets (e.g., code, data, models) or curating/releasing new assets...
\begin{enumerate}
  \item If your work uses existing assets, did you cite the creators?
    \answerYes{}
  \item Did you mention the license of the assets?
    \answerYes{}
  \item Did you include any new assets either in the supplemental material or as a URL?
    \answerYes{}
  \item Did you discuss whether and how consent was obtained from people whose data you're using/curating?
    \answerNo{}
  \item Did you discuss whether the data you are using/curating contains personally identifiable information or offensive content?
    \answerNo{}
\end{enumerate}

\item If you used crowdsourcing or conducted research with human subjects...
\begin{enumerate}
  \item Did you include the full text of instructions given to participants and screenshots, if applicable?
    \answerNA{}
  \item Did you describe any potential participant risks, with links to Institutional Review Board (IRB) approvals, if applicable?
    \answerNA{}
  \item Did you include the estimated hourly wage paid to participants and the total amount spent on participant compensation?
    \answerNA{}
\end{enumerate}

\end{enumerate}


\clearpage
\begin{appendices}
\section{Change of variable}

As described in \secref{sec:led:change_of_variable}, we need to keep track of the density of the input random variable as it passes through different layers of the generator.
In our architecture we used fully connected, convolutional, batch norm, dimensionality inflation, and an activation layer.
We describe in the following each of these layers, how to estimate the output density given the input density, and how to compute the determinant of the Jacobian. 

\subsection{Fully connected}
A fully connected layer can be represented as
\begin{equation}
    t = wu + b
    \label{eq:supmat:led:fully_connected}
\end{equation}
Here $u\in \vecspace{R}^{o_1}$ is the input variable, $w \in \vecspace{R}^{o_1 \times o_1}$ is a square weight matrix, and $b \in \vecspace{R}^{o_1}$ is the bias vector.
To compute the output density using the change of variable formula, we simply compute the log determinant of the weight matrix using PyTorch's \textit{slogdet()} function.
One must ensure $w$ to be full rank in order for the change of variable formula to still apply. However, since our objective is to maximize entropy of the generator and therefore maximize the absolute determinant, a good initialization ensures $w$ to have full rank.

\subsection{Activation}
\begin{figure}
    \centering
    \includegraphics[width=0.6\columnwidth]{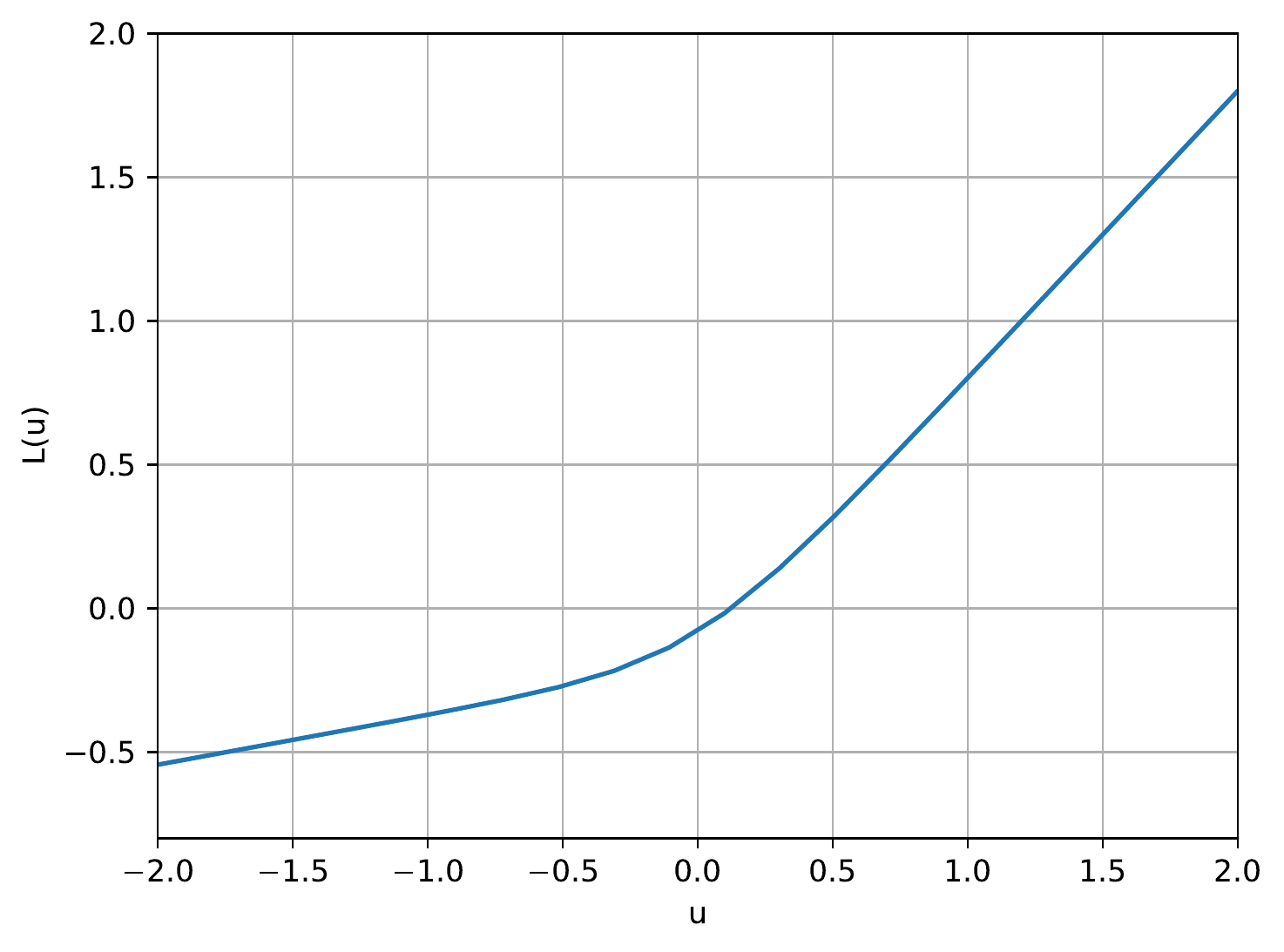}
    \caption{Our custom activation function as described in Eq. \ref{eq:custom_activation} with $\alpha=5.8$ and $\beta=1$.}
    \label{fig:custom_activation}
\end{figure}

Since ReLU is not an invertible operation, it is not suitable to use it in our generator.
However, several modern generator architectures use ReLU, and it has been shown to improve performance of neural networks empirically.
Therefore, we design our own activation function that closely follows ReLU and yet is suitable for our use case.
We define our custom activation as:
\begin{equation} \label{eq:custom_activation}
    L(u)=\frac{1}{\alpha}\log\left(e^{\alpha\left(u-u_{0}\right)}+e^{\beta\left(u-u_{0}\right)}\right)-t_{0}
\end{equation}
Where $\alpha$ and $\beta$ are user-defined and $u_0$ and $t_0$ are offsets.
We set $\alpha=5.8$ and $\beta=1$ (Fig. \ref{fig:custom_activation}).
This activation results in a constant slope of $1$ in the positive domain. In the negative domain, the slope and elbow can be controlled by adjusting $\alpha$ and $\beta$.
Since the activation applies to every component of its input vector independently, the Jacobian of this layer is a diagonal matrix.
In the case of diagonal matrices, the log determinant is easily computed as the sum of the logarithm of each element's derivative. 
The derivative of each element can be easily calculated as:
\begin{equation}
    \frac{\text{d}L\left(u\right)}{\text{d}u}=\frac{e^{\alpha\left(u-u_{0}\right)}+\frac{\beta}{\alpha}e^{\beta\left(u-u_{0}\right)}}{e^{\alpha\left(u-u_{0}\right)}+e^{\beta\left(u-u_{0}\right)}}
\end{equation}

\subsection{Convolution} 
The naive method to compute the determinant of the Jacobian of a convolutional layer would be to compute the determinant of the corresponding Toeplitz matrix of the convolution kernel.
As the layers keep increasing in size, so do the Toeplitz matrices, and computing their determinant becomes a slow task.
To avoid this, we use the method described by \cite{sedghi2018the} to compute the singular values and the determinant of the convolution.
Their algorithm uses the fact that the circulant matrices are diagonalized by the Fourier transform.
As a result, we are constrained to use convolution with circular padding.
With that small change, we can compute the log determinant quickly.

\subsection{Batch-norm}
We take this layer verbatim from \cite{dinh2016density} as 
\begin{equation}
    B\left(u\right)=\frac{u-\text{E}\left[u\right]}{\sqrt{\text{Var}\left[u\right]+\epsilon}}\times\gamma+\beta
\end{equation}
Note that $\text{E}\left[u\right]$ and $\text{Var}\left[u\right]$ are function of all elements in $u$.
Therefore, the Jacobian is not diagonal and is in general hard to compute.
In our current implementation, we disregarded the fact that  $\text{E}\left[u\right]$ and $\text{Var}\left[u\right]$ are functions of $u$ and treated them as constants.
As a results, the Jacobian is diagonal with each element being the simple derivative:
\begin{equation}
    \frac{\text{d}B\left(u\right)}{\text{d}u_{i}}=\frac{\gamma}{\sqrt{\text{Var}\left[u\right]+\epsilon}}
\end{equation}

%

\subsection{Dimensionality inflation}
As described in \secref{sec:led:change_of_variable}, we increase dimensionality by introducing noise and copying over the left neighbor in an image, emulating the nearest neighbor interpolation. One can also device bi-linear or bi-cubic interpolation methods. This, however, is left as future work.

\section{Implementation}
\subsection{Implementation Details} \label{sec:implemetation_dets}
We trained the generator and discriminator using the PyTorch ADAM optimizer~\cite{adam}.
The learning rate for both generator and discriminator was set to 1e-3. 
We trained LED on the MNIST~\cite{mnist} and CelebA~\cite{celeba} datasets for 500 epochs.
We saved the models after every epoch and presented the result of the epoch with the highest FID.

\subsection{Perpendicular convolution}
In \secref{sec:perp_cmp_in_prob_comp}, we described how one might help the density estimation function $\nn$ assign low density to off-the-image manifold points. 
The computation of the perpendicular input component in case of linear layer is shown in \secref{sec:perp_cmp_in_prob_comp}. 
When this linear operation 
is implemented using convolutions, one can perform transposed convolution to compute $t^T*w$, where $t$ and $w$ are as defined in \secref{sec:perp_cmp_in_prob_comp}.
Importantly, however, we still have to `lift' the dimensionality of the generated samples to the dimension of the input data dimension (number of pixels in case of images).
The reason behind this 'lifting' is that data embedded in a higher-dimensional manifold is a curve without width. 
Having no width essentially means that the density function is a Dirac delta function over this curve
Since the the Dirac delta function has no gradients outside of the embedded curve, learning such a function using gradient descent is ill-posed.

\subsection{Generator: Random sampling from a Gaussian}
During training, we set the generator in such a way that for each sample in a batch, with some probability, the values of its corresponding output will be replaced by samples from a normal distribution.
Adding this Normal sampling defines a non-zero probability distribution over the codomain of the generator, since for each point the generator cannot generate on its own, the normal distribution has a non-zero probability.

\subsection{Numerical stability in generator loss}
In our experiments we found that the two terms in Eq. \ref{kl_rejects_normalization} have different orders of magnitude.
This difference prevented the optimization to converge.
We solved this disparity by adding a weight coefficient $k$ to the entropy term $H(P_G(\rv{y}))$ to scale it down.
In practice, this changes our KL-divergence into $f$-divergence with $f\left(t\right)=t\log\left(t^{k}\right)$.
In comparison, KL-divergence is a special case of  $f$-divergence with $f\left(t\right)=t\log\left(t\right)$.

Looking again at Eq. \ref{kl_rejects_normalization} with the new divergence formulation, $N_{\theta}\left(y_{i}\right)$ should be reformulated as $N_{\theta}^{k}\left(y_{i}\right)$.
Instead, we redefine our discriminator's output to be $k\log N_{\theta}\left(y_{i}\right)$.
With this redefinition we maintain the stability of the generator loss while not affecting the discriminator loss, since $k$ will be implicitly computed in the integral approximation of Eq. \ref{int_as_expectation}.

\section{Architecture}
\label{LED:sec:architecture}

\subsection{DC-GAN based} \label{sec:dcgan_arch}
The generator architecture we used for MNIST and CelebA is based on the DC-GAN architecture.
Given a latent vector, the generator first increases the dimensionality by using a linear layer which is then padded with noise.
The size of the hidden latent image is then doubled on the expense of the number of channels multiple times, until the number of channels reaches the target number (1 for MNIST, 3 for CelebA).
The image is then down-sampled (\secref{LED:sec:dim_red}) until it reaches the required number of pixels.

\subsection{Expressive power of the generator for a toy problem} \label{sec:toy_arch}
For the 2D synthetic data (\secref{sec:results_synth_data}) we follow the same principles as in \secref{sec:dcgan_arch}, but with slight differences.
The generator receives as input a 2D point.
This 2D point is expanded to have the maximal number of features by the use of linear layers, where each linear layer doubles the number of features. 
The maximal number of features is defined by the user before initialization
After the expansions, the resulting latent vector goes through a number of blocks, where each block contains a linear layer that maintains dimensionality, a batch normalization layer and an activation function.
The number of blocks is defined by the user before initialization.
For the results shown in the paper we used 256 features and 8 blocks.  
The result is then down-sampled (\secref{LED:sec:dim_red}) until it is 2D.

In \secref{sec:results_synth_data} we showed that our generator could not capture all the modes. 
We believe that this was caused by the fact that the generator architecture was not expressive enough for the task.
To support this, the effects of different maximal number of features and number of blocks are shown in \secref{sec:gen_power_2d}. 
We can see that with a smaller number of blocks the generator can capture more modes.
Theoretically, with a large number of blocks the generator could capture more modes, albeit with very long computation times.
In the future we will look into ways to increase the expressive power of the generator without excessively increasing the computational time.

\section{Approximation quality of the integral}
\begin{figure}
    \centering
    \includegraphics[width=\columnwidth]{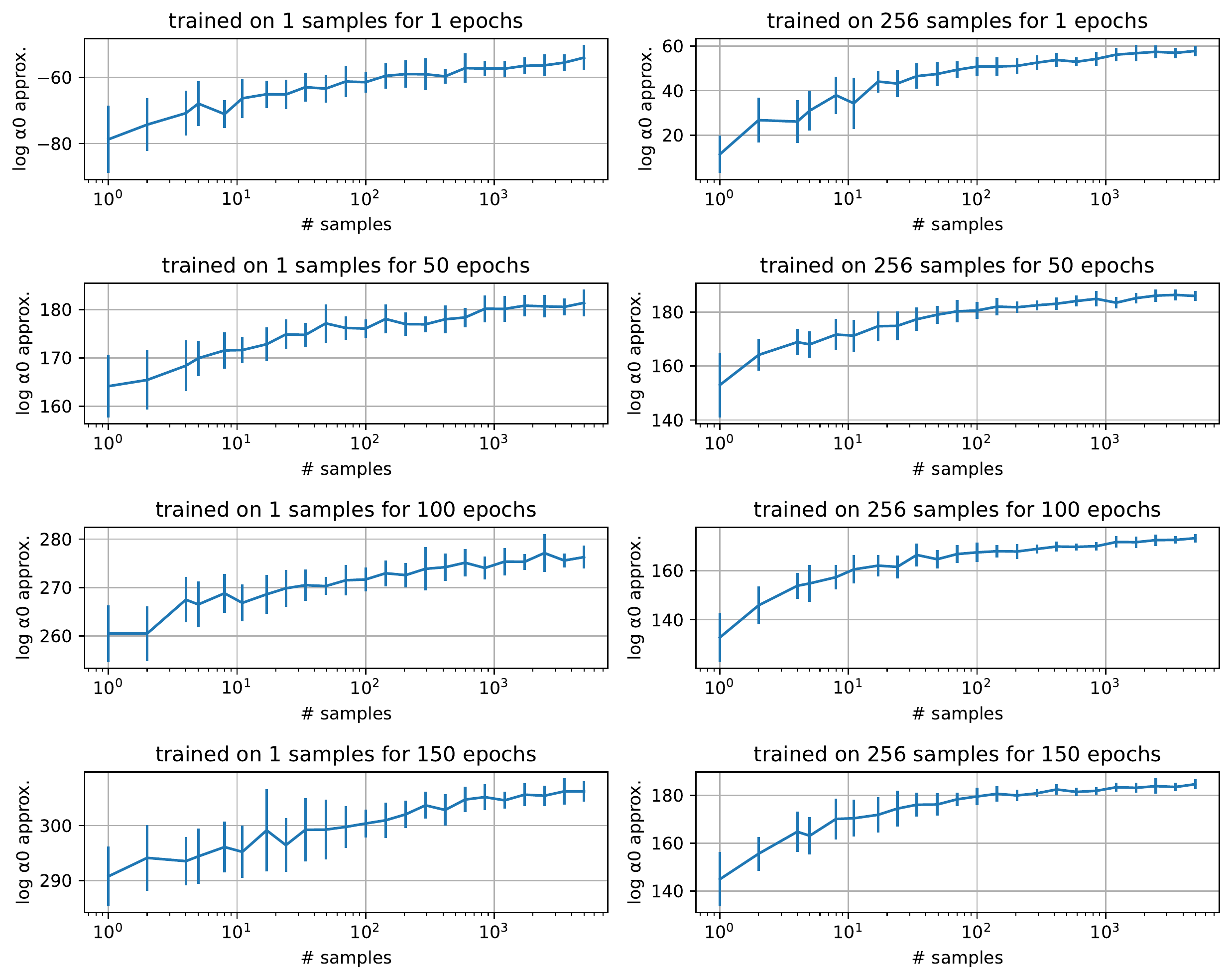}
    \caption{The approximated integral as a function of the number of samples. The left and right columns show the results of models that were trained with 1 and 256 samples, respectively, for the integration approximation. Each row corresponds to a different number of epoch. The $y$-axis of each graph is the approximated integral value and the $x$-axis is the number of samples used to approximate the integral post-training. The error bar correspond to $1\sigma$.}
    \label{fig:alpha_approx}
\end{figure}

In order to see how many samples are required and how the training time affects the approximation, we compared the approximated $\alpha_\theta$ across two models, one trained with 1 sample and the other with 256 samples. After training each model for a certain number of epochs, we approximated $\alpha_\theta$ with a different number of samples. We calculated $\alpha_\theta$ 20 times for each model and epoch, and report the means and standard deviations in Fig. \ref{fig:alpha_approx}.
Fig. \ref{fig:alpha_approx} shows us that when training the model with 256 samples (left column), $\alpha_\theta$ converges to a constant value across epochs (about $185$), with the standard deviation decreasing when using more samples.
This means that our model is consistent and captures the domain well.
On the other hand, when using 1 sample, the approximated integral does not saturate and has a different range every epoch.
It also shows us that the standard deviations do not decrease even when using more samples.
Fig. \ref{fig:alpha_approx} also shows us that after a few epochs when using a single sample, the approximated integral is very small. This in turn results in a high likelihood which explains the high $b/p$ value for 1sa in Tbl. \ref{table:FID_score_b_ps}.


\clearpage

\section{Additional results}

\subsection{Generator expressiveness and power} \label{sec:gen_power_2d}
\begin{figure}[h]
    \centering
    \includegraphics[width=0.84\columnwidth]{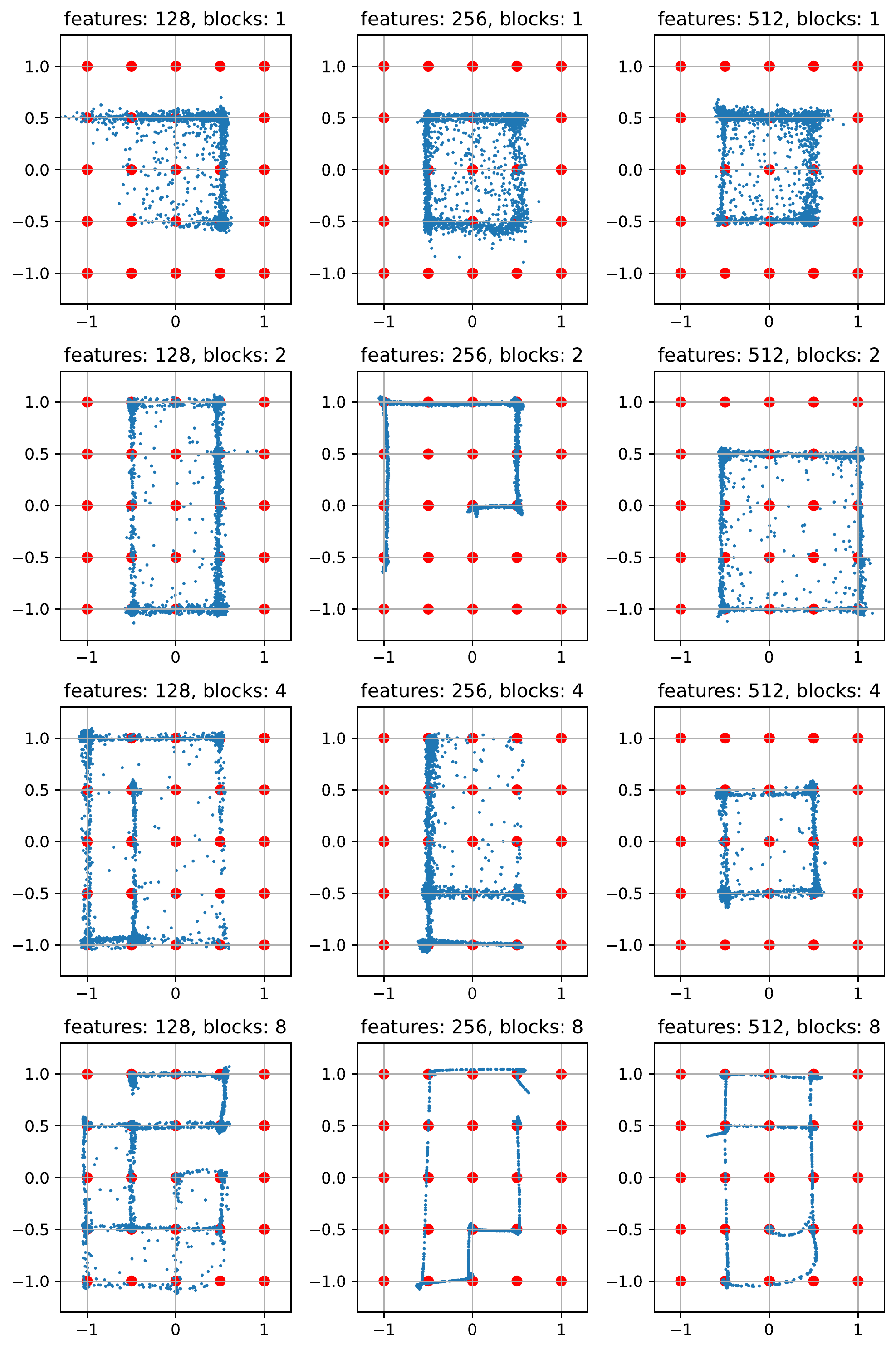}
    \caption{Randomly generated 2D points from generators with different numbers of layers. The generators were trained on the grid of GMMs (Fig. \ref{fig:toy_dists}g). The red dots represent the true modes and the blue dots are the generated data. Each column corresponds to a different size of latent space (number of features) and each row corresponds to a different number of blocks, as described in \secref{sec:toy_arch}.}
    \label{fig:gen2d_power}
\end{figure}

\newpage
\subsection{Generated images sorted by their probability} \label{sec:sort_prob_res}
\begin{figure}[h]
    \centering
    \includegraphics[width=\columnwidth]{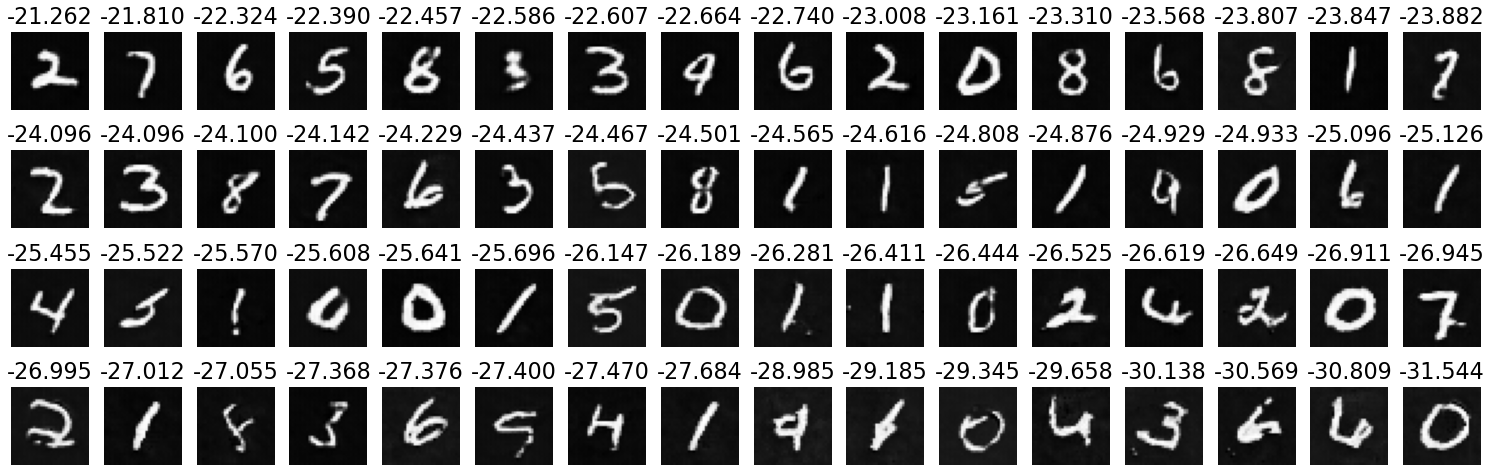}
    \caption{Generated MNIST images, sorted from high probability (top left) to low probability (bottom right). The number above each image is its unnormalized log-likelihood. Note that as the image quality drops, so does the probability.}
    \label{fig:mnist_prob_sort}
\end{figure}

\begin{figure}
    \centering
    \includegraphics[width=\columnwidth]{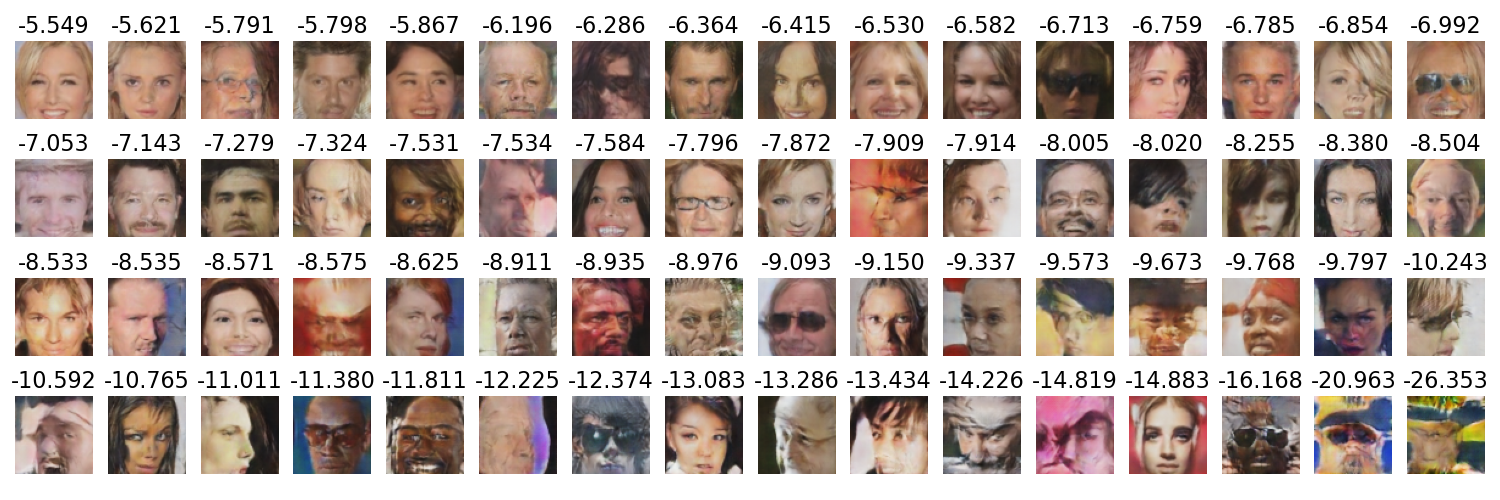}
    \caption{Generated CelebA images, sorted from high probability (top left) to low probability (bottom right). The number above each image is its unnormalized log-likelihood. Note that as the faces are more deformed, so their probability drops.}
    \label{fig:celeb_sort_prob}
\end{figure}

\newpage

\subsection{Interpolating the latent space for MNIST} \label{sec:interp_mnist}
\begin{figure}[h]
    \centering
    \includegraphics[width=\columnwidth]{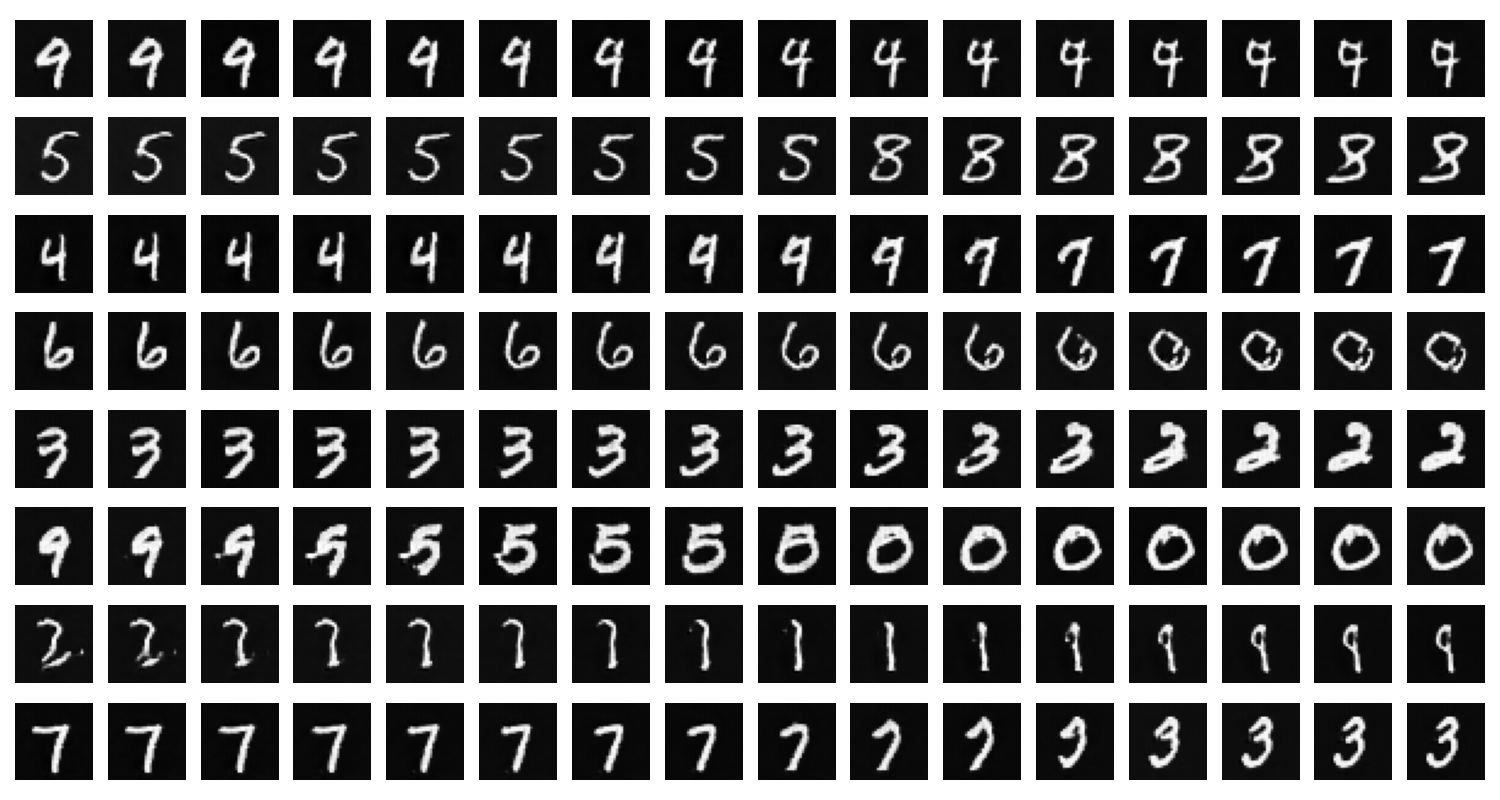}
    \includegraphics[width=\columnwidth]{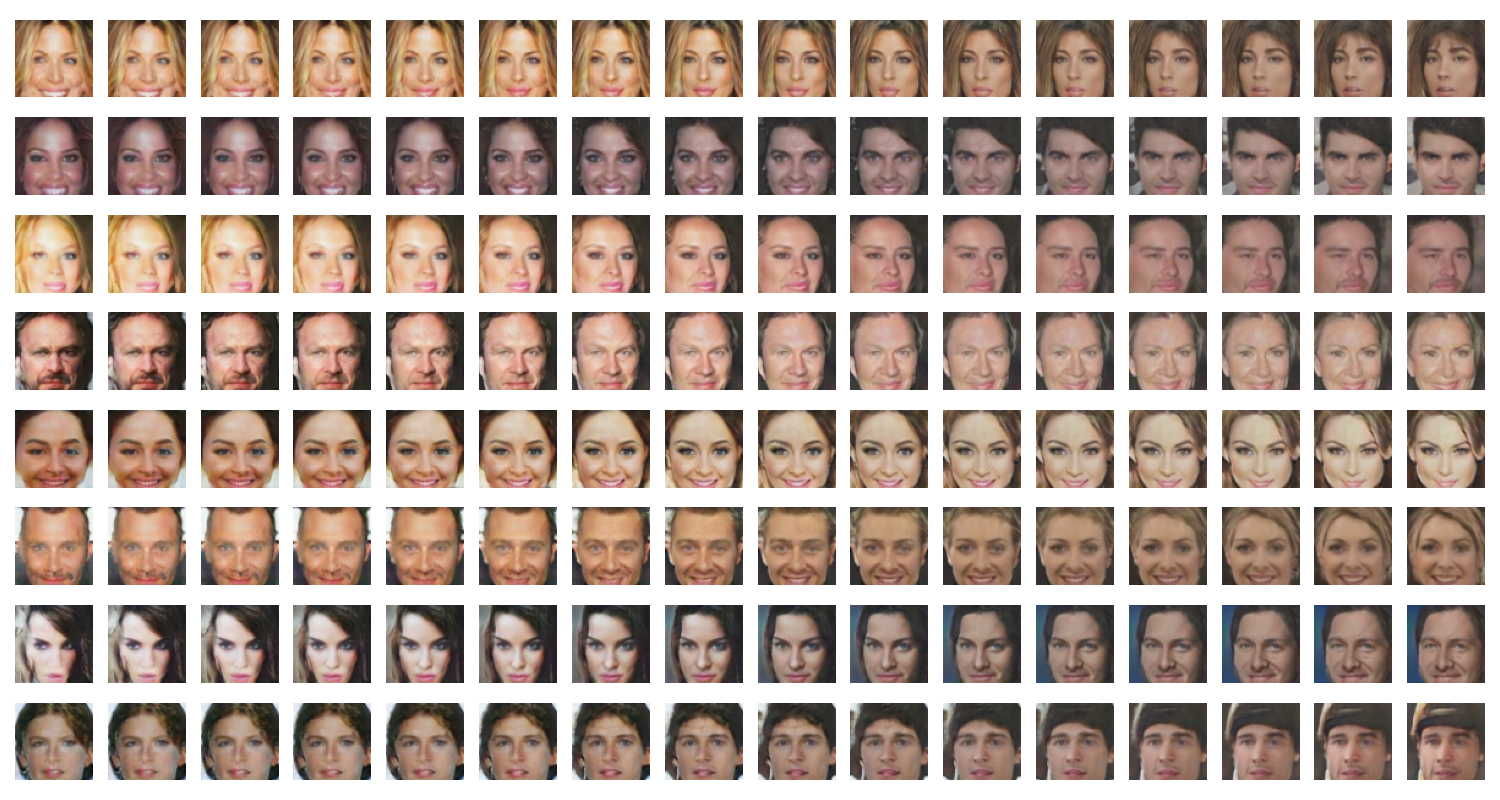}
    \caption{Interpolation over latent space. Each row corresponds to a different interpolation. The leftmost and rightmost columns are images generated by random latent vectors. The images between them are generated by the linear interpolations between the leftmost and rightmost columns.}
    \label{fig:mnist_interp}
\end{figure}

\newpage
\subsection{Optimizing the latent vector of far away points}\label{sec:mnsit_opt_far}
\begin{figure}[h]
    \centering
    \includegraphics[height=0.75\textheight]{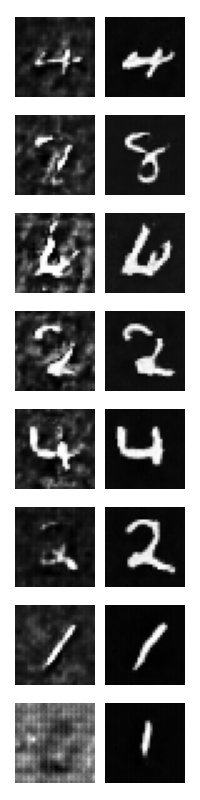}
    \includegraphics[height=0.75\textheight]{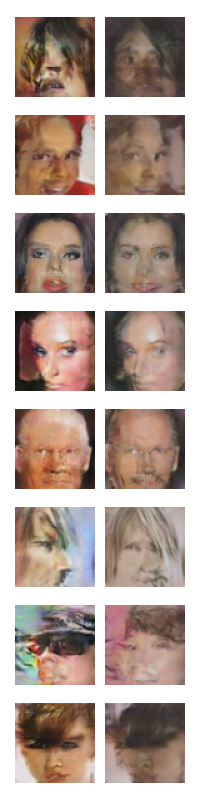}
    \caption{The result of maximizing the discriminator probability by optimizing the latent space. Left column: images generated by the initial latent vectors that are far from the mean ($4\sigma$). Left column: the corresponding generated images of the optimized latent vectors. Note that the optimization de-noised some of the background and filled missing holes.}
    \label{fig:mnist_opt_far}
\end{figure}
\end{appendices}

\end{document}